\documentclass[10pt,twocolumn,letterpaper]{article}

\usepackage{cvpr}              

\usepackage[table]{xcolor}
\usepackage{dashbox}
\usepackage[framemethod=tikz]{mdframed}
\usepackage{graphicx}
\usepackage{rotating}
\usepackage{wrapfig}
\newmdenv[hidealllines=false,roundcorner=5pt]{linebox}
\usepackage{multirow}
\usepackage{pifont}

\newcommand{\cmark}{\ding{51}}
\newcommand{\xmark}{\ding{55}}

\definecolor{cvprblue}{rgb}{0.21,0.49,0.74}
\usepackage[pagebackref,breaklinks,colorlinks,allcolors=cvprblue]{hyperref}

\def\paperID{10947} 
\def\confName{CVPR}
\def\confYear{2025}

\title{SFDM: Robust Decomposition of Geometry and Reflectance for Realistic Face Rendering from Sparse-view Images}

\author{Daisheng Jin$^{1}$ \quad Jiangbei Hu$^{1,2}$ \quad Baixin Xu$^{1}$ \quad Yuxin Dai$^{1}$ \quad Chen Qian$^{3}$ \quad Ying He$^{1}$\thanks{Corresponding author: Y. He (Email: yhe@ntu.edu.sg)}    \\
$^{1}$S-Lab, Nanyang Technological University, 
$^{2}$Dalian University of Technology, 
$^{3}$SenseTime Research\\
}

\begin{document}
\maketitle
\begin{abstract}



In this study, we introduce a novel two-stage technique for decomposing and reconstructing facial features from sparse-view images, a task made challenging by the unique geometry and complex skin reflectance of each individual. To synthesize 3D facial models more realistically, we endeavor to decouple key facial attributes from the RGB color, including geometry, diffuse reflectance, and specular reflectance. Specifically, we design a Sparse-view Face Decomposition Model (\textbf{SFDM}): \textbf{1)} In the first stage, we create a general facial template from a wide array of individual faces, encapsulating essential geometric and reflectance characteristics. \textbf{2)} Guided by this template, we refine a specific facial model for each individual in the second stage, considering the interaction between geometry and reflectance, as well as the effects of subsurface scattering on the skin. With these advances, our method can reconstruct high-quality facial representations from as few as three images. The comprehensive evaluation and comparison reveal that our approach outperforms existing methods by effectively disentangling geometric and reflectance components, significantly enhancing the quality of synthesized novel views, and paving the way for applications in facial relighting and reflectance editing. Visit our project page for more details \href{https://kingjg.github.io/SFDM.github.io/}{https://kingjg.github.io/SFDM.github.io/}.

\end{abstract}    
\vspace{-10pt}

\section{Introduction}
\label{sec:intro}

3D human face reconstruction finds widespread applications in various fields, including game design and film production. Developments in neural rendering techniques have shown promising performance and advantages in synthesizing photo-realistic images for complex scenes~\cite{mildenhall2021nerf,barron2021mip,muller2022instant,tancik2023nerfstudio}. Concurrently, a surge of research focusing on head and facial attribute reconstruction from images via implicit neural representations~\cite{gafni2021dynamic,park2021nerfies,park2021hypernerf,zheng2022avatar} has emerged. 
These innovative approaches address the limitations associated with traditional explicit methods, such as those based on the 3D Morphable Model (3DMM)~\cite{blanz2023morphable}, which struggled with issues like dependency on mesh resolution and difficulties in managing topological variations.
Furthermore, implicit neural representation methods demonstrate their effectiveness in 3D face reconstruction from low-view images~\cite{xu2023deformable}, expression transfer~\cite{athar2022rignerf,hong2022headnerf,lin2023single}, and animations~\cite{zheng2022structured,zheng2022avatar,giebenhain2023learning}.

\begin{figure}[!t]
    \centering \includegraphics[width=1.0\linewidth]{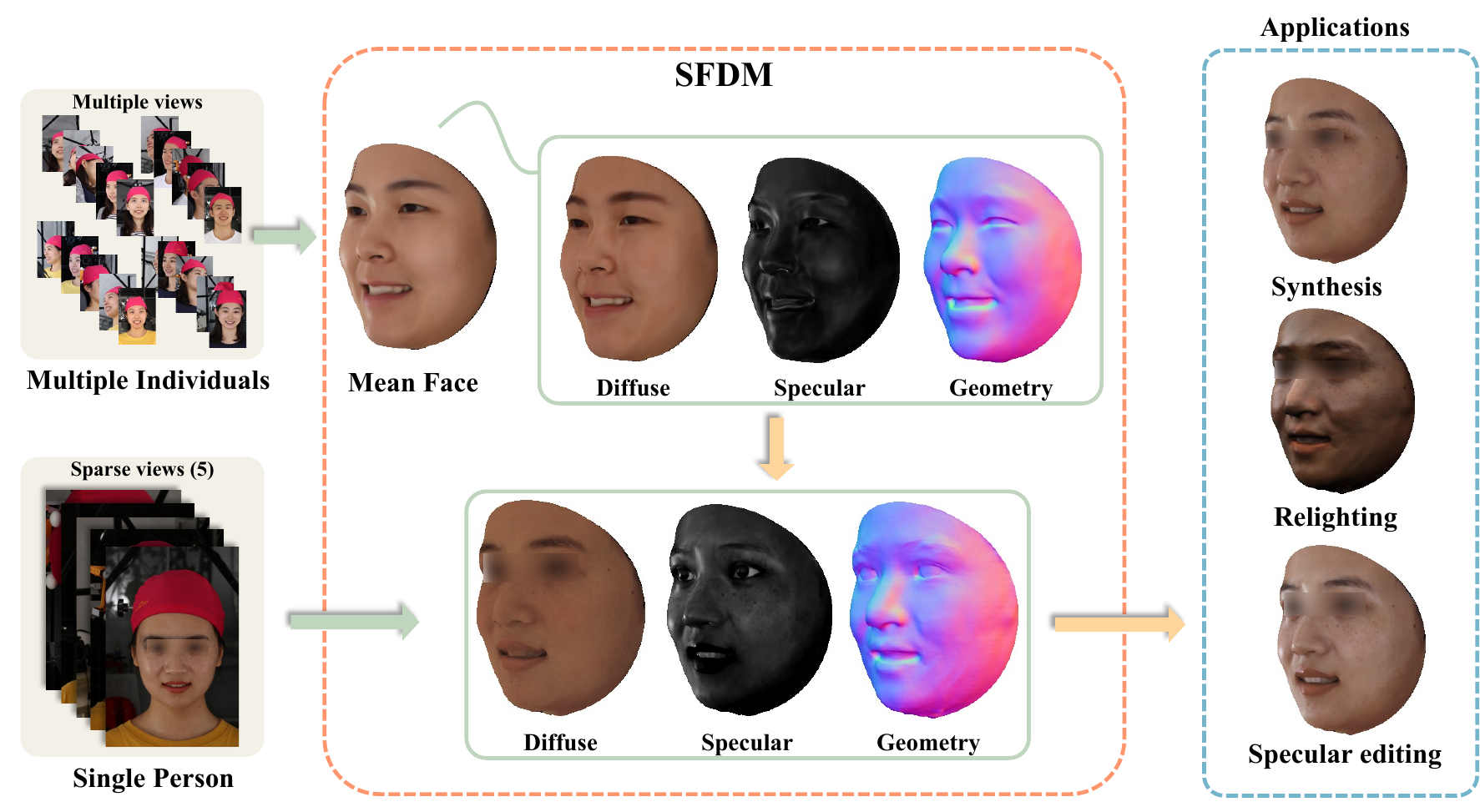}
    \vspace{-10pt}
        \caption{Our method learns a general facial template with distinct key facial attributes from multiple individuals. This template enables photorealistic novel-view synthesis for each individual, as well as other applications.}
    \label{fig:template}
    \vspace{-15pt}
\end{figure}

In 3D face model rendering, both accurate shape attributes and high-quality reflectance properties are essential for synthesizing photorealistic faces. Thus, robustly decomposing key facial attributes including geometry, diffuse and specular reflectance, and environment light is a crucial task, which not only enhances the accuracy of face reconstruction but also opens up possibilities for more applications like relighting~\cite{zhang2021nerfactor,boss2020two,li2020inverse,chen2022relighting4d,he2024diffrelight}.
However, most previous studies only use neural networks to learn radiance and geometric information from images, neglecting the effect of rendering-related physical information such as facial skin characteristics and environmental lighting conditions. NeRV~\cite{srinivasan2021nerv}, PhySG~\cite{zhang2021physg}, and following methods~\cite{boss2021nerd,munkberg2022extracting,boss2021neural,cheng2021multi,hasselgren2022shape,liu2023nero} incorporate the Bidirectional Reflectance Distribution Functions (BRDFs) into the neural rendering framework, facilitating the decomposition and joint reconstruction of shape, reflectance, and illumination.  Nonetheless, these methods primarily concentrate on non-biological materials,
which are not suitable for decomposing face models that possess complex attributes.
NeuFace~\cite{zheng2023neuface} introduces an approximated BRDF integration and learns the physically meaningful underlying representations for 3D face reconstruction. Nevertheless, due to the ill-posedness of decomposition, NeuFace often converges to a suboptimal solution under sparse-view settings, in which specular reflectance is absent, and all radiance is derived from diffuse attributes. In addition, the accuracy of geometric reconstruction significantly diminishes as the number of views decreases. Therefore, it still poses a significant challenge to decompose shape, reflectance, and environment light attributes from sparse view images in 3D face model reconstruction.

In this study, we introduce a novel two-stage face decomposition framework, termed Sparse-view Face Decomposition Model (\textbf{SFDM}), aimed at reconstructing 3D faces from sparse-view images. The pipeline of our proposed algorithm is depicted in Fig.~\ref{fig:pipeline}. Based on the observation that human faces share similar geometric features and reflectance properties, we first derive a novel general facial template from images of a set of varied individual faces in Stage 1. In Stage 2, we employ the learned facial template with both geometry and reflectance attributes to refine the decomposition of each individual face model. In both stages, we decompose the shape and material attributes using a geometry module and a reflectance module, respectively. 

Specifically, the geometry module consists of a template network and a deformation network. In the first stage, we can get a template model with common geometry features based on the training of multi-view images from varied faces, which can be further transformed to a specific individual face model by deformation. 
Building on the facial geometry template, we innovatively propose a reflectance module that integrates reflectance information into the facial template. Specifically, we introduce the BRDF as a coarse representation of reflectance for facial skin. We employ a BRDF template network to learn the shared characteristics of diffuse and specular reflectance properties, and a BRDF offset network is utilized to reconstruct identity-specific reflectance details. Furthermore, we propose a novel albedo gradient predictor designed to capture more general facial reflectance attributes.
In the second stage, based on the learned template with geometry and reflectance attributes, we refine the attribute details from sparse-view images of each individual face model. 
To this end, we incorporate a displacement network and a reflectance offset network on the learned template, employing a design that harmonizes geometry and reflectance information to capture high-frequency details of facial attributes.
In addition, we introduce a diffuse offset network to take into account the effect of subsurface scattering, thus providing a more realistic simulation of the facial skin.
We summarize the main contributions of this study as follows:
\begin{itemize}
    \item We propose a novel decomposition framework for human facial representations to learn a template with various attributes, including geometric features, diffuse reflectance, and specular reflectance.
    \item Based on the learned template, our tailored refinement stage enables robust decomposition of attributes for each individual face under as few as three images, attaining superior geometric accuracy and reflectance detail.
    \item Extensive experiments demonstrate the face decomposition results obtained through our framework can facilitate more realistic synthesis in novel views, as well as applications such as specular editing and relighting.
\end{itemize}


\section{Related work}
\label{sec:related}



\textbf{Physically based rendering.} To achieve a photorealistic rendering effect, physically based rendering (PBR)~\cite{pharr2023physically} is proposed to simulate the light and surfaces using real-world optical principles. Reversely, physically based inverse rendering seeks a way to decompose light and surface factors from the images~\cite{bi2020neural,munkberg2022extracting,bi2020deep,boss2021nerd}. IDR~\cite{yariv2020multiview} proposes a surface reconstruction method that simultaneously disentangles the geometry and appearance. Subsequent works incorporate inverse rendering into the neural rendering pipeline, developing various methods for general objects~\cite{zhang2021physg,boss2021neural}, and enhancing decomposition effects for objects with specialized materials~\cite{liu2023nero,che2020towards}.
NeuFace~\cite{zheng2023neuface} proposes a decomposition pipeline for human faces, taking into account the sophisticated properties of facial skin reflectance. 
However, without ground truth for reflectance components, face decomposition is an ill-posed problem, causing NeuFace to often struggle with accurately decomposing reflectance. Additionally, NeuFace does not consider the effect of subsurface scattering on facial skin, resulting in less realistic PBR~\cite{yang2023light}. In this work, with our novel facial template and the single-person refinement module, which further considers subsurface scattering effects, we can robustly decompose geometry and reflectance in a more realistic manner from sparse-view images.

\noindent\textbf{Face reconstruction.} Given the inherent similarities among human faces, it is intuitive to develop a template that serves as a prior to enhance facial reconstruction. Based on a large amount of facial data, 3DMM~\cite{blanz2023morphable} builds a 3D morphable face to describe the transformation of shape and expression, which is widely used in face recognition~\cite{zhao2003face}, 3D reconstruction, face editing~\cite{park2021nerfies,feng2021learning}, etc.
Using 3D morphable faces also enables face reconstruction from monocular images~\cite{zollhofer2018state,feng2022towards, dib2024mosar, papantoniou2023relightify, lattas2023fitme}. For instance, Han \textit{et al.}~\cite{han2023learning} build a 3D morphable face with reflectance on BFM09~\cite{paysan20093d} for material editing. 
However, in face reconstruction, the use of 3D morphable faces often leads to a loss of detail during the compression and fitting process, resulting in overly coarse reconstructions. Moreover, given limited view information, monocular methods typically require reflectance ground truth for training~\cite{smith2020morphable}.
As implicit neural representation shows extraordinary efficacy for 3D representation, numerous studies~\cite{yenamandra2021i3dmm,zheng2022imface,zheng2022avatar,xu2023deformable} construct 3D parametric face or head models based on neural networks, yielding improved geometric details for faces or heads. 
Besides, some facial capturing methods utilize more informative inputs such as videos~\cite{wang2023sunstage} and polarized images~\cite{azinovic2023high}, enabling high-quality reconstruction across a wider range of scenarios.
Our goal is to achieve face reconstruction solely from sparse-view images, without any reflectance ground truth.
Our Stage 1 employs an implicit representation to construct a facial template, incorporating prior knowledge of facial geometry and reflectance. This approach enables robust decomposition and realistic face reconstruction during the single-person refinement process under sparse views.

\section{Preliminary}
The effectiveness of PBR has been validated by numerous studies~\cite{kajiya1986rendering,verbin2022ref,zheng2023neuface} for photo-realistic rendering in scenarios with complex reflection properties.
Inspired by these works and as illustrated in Fig.~\ref{fig:render_process}, we formulate the output radiance $L_{\text{o}}\in\mathbb{R}^3$ at surface position $\mathbf{x}\in\mathbb{R}^3$ along output direction 
$\boldsymbol{\omega}_{\text{o}}\in\mathbb{R}^3$ as:
$L_{\text{o}}(\mathbf{x},\boldsymbol{\omega}_{\text{o}}) =\int_{\Omega }f(\mathbf{x},\boldsymbol{\omega}_{\text{i}},\boldsymbol{\omega}_{\text{o}}) L_{\text{i}}(\mathbf{x}, \boldsymbol{\omega}_{\text{i}})(\boldsymbol{\omega}_{\text{i}}\cdot \mathbf{n})\,\,\text{d}\boldsymbol{\omega}_{\text{i}},$
where $f$ represents the BRDF evaluation, $L_{\text{i}}\in\mathbb{R}^3$ is the incident light intensity from input direction $\boldsymbol{\omega}_{\text{i}}\in\mathbb{R}^3$, $\mathbf{n}\in\mathbb{R}^3$ is the surface normal. To decouple the view-dependent and view-independent radiance, we decompose $L_{\text{o}}$ into the diffuse term and the specular term by:
\begin{equation*}
\begin{split}
&L_{\text{o}}(\mathbf{x},\boldsymbol{\omega} _{\text{o}})=\underset{\text{diffuse}}{\underbrace{\frac{b_{\text{a}}(\mathbf{x})}{\pi} \int_{\Omega }L_{\text{i}}(\mathbf{x},\boldsymbol{\omega}_{\text{i}})(\boldsymbol{\omega}_{\text{i}}\cdot\mathbf{n})\,\, \text{d}\boldsymbol{\omega}_{\text{i}}}} \\
&+\underset{\text{specular}}{\underbrace{\int_{\Omega}f_{s}(\mathbf{x},\boldsymbol{\omega}_{\text{i}},\boldsymbol{\omega}_{\text{o}};b_{\text{s}}(\mathbf{x}),b_{\text{r}}(\mathbf{x}))L_{\text{i}}(\mathbf{x},\boldsymbol{\omega}_{\text{i}})(\boldsymbol{\omega}_{\text{i}}\cdot\mathbf{n})\,\, \text{d}\boldsymbol{\omega}_{\text{i}}}},
\end{split}
\label{eq:decompose}
\end{equation*}
where $b_{\text{a}}(\mathbf{x})\in\mathbb{R}^3$, $b_{\text{s}}(\mathbf{x})\in\mathbb{R}^4$, $b_{\text{r}}(\mathbf{x})\in\mathbb{R}$ are diffuse albedo, specular, and roughness fields obtained from BRDF, respectively. $f_s$ represents the BRDF evaluation that only accounts for the specular property. Furthermore, we improve the efficiency of integration computation by using the approximated pre-computed method~\cite{boss2021neural} as: 
\begin{equation}\label{eq:int_L_o}
    L_{\text{o}}(\mathbf{x}, \boldsymbol{\omega}_{\text{o}}) 
    \approx \underbrace{\frac{b_{\text{a}}(\mathbf{x})}{\pi } \hat{L}_{\text{i}}(\mathbf{n}, 1)}_{\text{diffuse}}
    +\underbrace{\hat{f}_{\text{s}}(\mathbf{x};b_{\text{s}}(\mathbf{x}),b_{\text{r}}(\mathbf{x}))\hat{L}_{\text{i}}\left(\boldsymbol{\omega}_{\text{r}}, b_{\text{r}}(\mathbf{x})\right)}_{\text{specular}},
\end{equation}
where $\boldsymbol{\omega}_{\text{r}}\in\mathbb{R}^3$ represent the mirrored view direction, $\hat{L}_{i}$ is the integration of input light, $\hat{f}_{s}$ is the integration of specular reflectance. Specifically, following ~\cite{karis2013real}, we employ an MLP to simulate two look-up-textures (LUT) $B_0$ and $B_1$ to calculate the BRDF integration as:
\begin{equation}
\hat{f}_{\text{s}}(\mathbf{x})=b_{\text{s}}(\mathbf{x})[F_{0}(\boldsymbol{\omega}_{\text{o}}, \mathbf{n}) B_{0}(\boldsymbol{\omega}_{\text{o}}\cdot \mathbf{n}, b_{\text{r}}(\mathbf{x})) +B_{1}(\boldsymbol{\omega}_{\text{o}}\cdot\mathbf{n}, b_{\text{r}}(\mathbf{x}))],
\end{equation}
where $F_{0}$ is the Fresnel term at normal incidence as ~\cite{boss2021neural}. For light representation, we use spherical harmonics~\cite{tunwattanapong2013acquiring} to approximate environment light by:
\begin{equation}
    L_{\text{i}}\left(\omega_{\text{i}}\right) \approx \sum_{\ell=0}^{k} \sum_{m=-\ell}^{\ell} c_{\ell }^{m} Y_{\ell }^{m}\left(\omega_{\text{i}}\right),
    \label{eq:sh}
\end{equation}
where $Y$ represents spherical basis functions and $c$ denotes corresponding coefficients. Considering that the facial material is not of high specular reflectance, we set the order as $k=10$~\cite{chen2020neural}. To integrate incident light rays, the light for the specular term can be integrated to the mirrored direction of the observation view, denoted as $\hat{L}_{\text{i}}\left(\boldsymbol{\omega}_{\text{r}}, b_{\text{r}}\right)$, and light for the diffuse term (Lambertian) can be approximated by $\hat{L}_{\text{i}}\left(\mathbf{n}, 1\right)$~\cite{boss2021neural}, which assumes light in the normal direction hits a completely rough surface. For our SH light representation, we integrate $\hat{L}_{\text{i}}\left(\boldsymbol{\omega}_{\text{r}}, b_{\text{r}}\right)$ for the specular term and $\hat{L}_{\text{i}}\left(\mathbf{n}, 1\right)$ for the diffuse term through an approximation using the first ten orders of spherical harmonics as ~\cite{zheng2023neuface}.

\begin{figure*}[!t]
    \centering
    \includegraphics[width=1.0\linewidth]{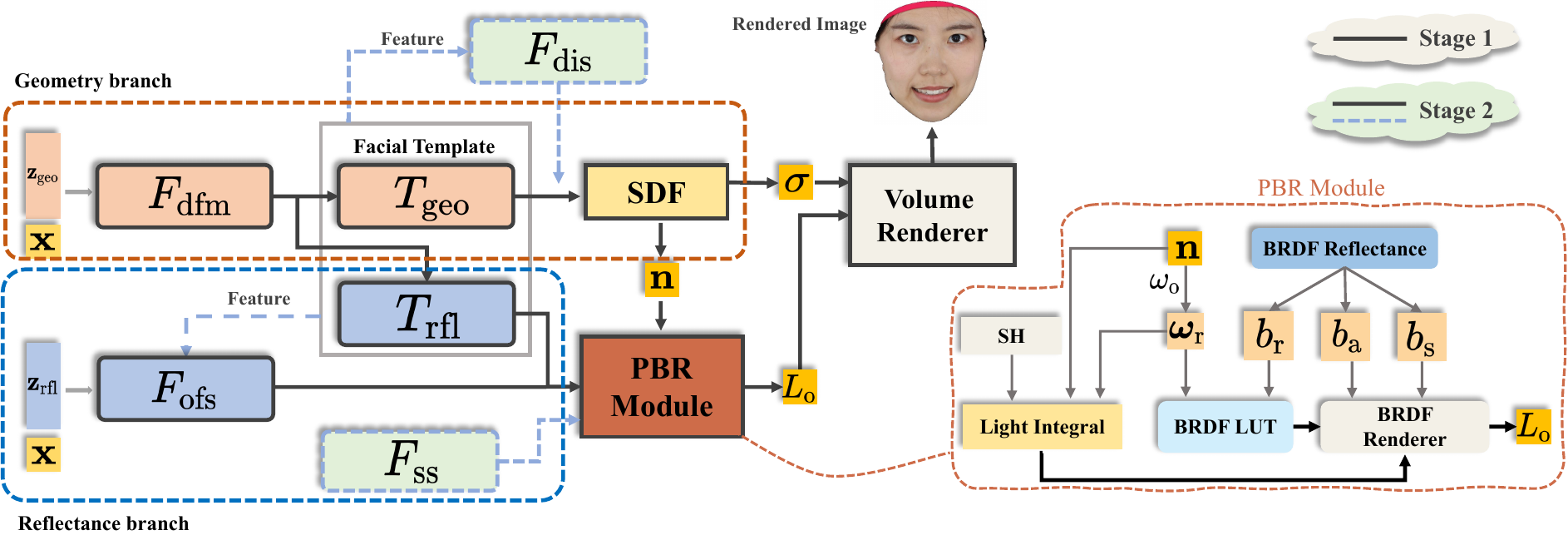}
    \vspace{-10pt}
    \caption{\textbf{Pipeline of SFDM.} We propose a geometry and reflectance decomposition framework for 3D face reconstruction from multi-view images. We first learn a general facial template from multi-view images of multiple individuals in Stage 1. Then the learned facial template guides the further attribute refinement for each individual face model in Stage 2. }
    \label{fig:pipeline}
    \vspace{-10pt}
\end{figure*}

\section{Method}
\label{sec:method}
Typical human faces share common features, such as two eyes and one nose, with variations in characteristics such as shape, skin color, and skin texture. To better utilize the shared features of human faces, we divide the decomposition process into two stages to learn the neural fields of human faces from multi-view images. Fig.~\ref{fig:pipeline} presents the architecture of the Sparse-view Face Decomposition Model (\textbf{SFDM}). In Stage 1, we create a general facial template with geometry and reflectance attributes from a collection of portraits of multiple individuals. In Stage 2, we refine the specific attributes for each individual face model based on the learned facial template under sparse-view settings.


\subsection{Generating facial template}
\label{subsec:stage1}
Given a collection of portraits of multiple individuals $\{\mathcal{I}_i\}$, we first learn a general facial template with both geometry and reflectance attributes.

\noindent\textbf{Geometry template.} Following the previous work~\cite{xu2023deformable,zheng2022imface}, we jointly employ a geometry template network $T_{\text{geo}}$ and a deformation network $F_{\text{dfm}}$ to reconstruct the geometry (represented by signed distance field $\mathcal{S}$) from given images. As illustrated in Fig.~\ref{fig:pipeline}, we formulate the geometry network as:
\begin{equation}\label{eq:g-temp}
    \mathcal{S}(\mathbf{x})=T_{\text{geo}}(F_{\text{dfm}}(\mathbf{x}, \mathbf{z}_{\text{geo}})),
\end{equation}
where $\mathbf{x}\in\mathbb{R}^3$ is a query point, $\mathcal{S}(\mathbf{x})\in\mathbb{R}$ returns the signed distance value. We take $\mathbf{z}_{\text{geo}}\in\mathbb{R}^{128}$ as an identity-related latent code that encodes geometry, so we can get the unique SDF for each individual. Similar to VolSDF~\cite{yariv2021volume}, we convert the SDF value to density $\sigma$ and then feed it into the volume rendering block.

\noindent\textbf{Reflectance template.} 
Apart from geometry attributes, we also consider reflectance attributes, which are equally crucial and share similar features across human faces. Therefore, we aim to create a coarse but general template that captures the underlying common reflectance characteristics of human faces in Stage 1.

To obtain the reflectance attributes, we approximate the Bidirectional Reflectance Distribution Function (BRDF) by the network that consists of a reflectance template $T_{\text{rfl}}$ and a BRDF offset network $F_{\text{ofs}}$. Similar to the geometry template, we leverage $T_{\text{rfl}}$ to fit the general facial BRDF for multiple individuals and then offset the BRDF value of the query points on each face model through $F_{\text{ofs}}$. We formulate the calculation of BRDF $\mathcal{B}_{\text{st1}}(\mathbf{x})$ as follows:
\begin{equation}
    \mathcal{B}_{\text{st1}}(\mathbf{x})=T_{\text{rfl}}[F_{\text{dfm}}(\mathbf{x}, \mathbf{z}_{\text{geo}})] + wF_{\text{ofs}}(\mathbf{x}, \mathbf{z}_{\text{rfl}}),
\end{equation}
where the query points $\mathbf{x}$ in the observation space are deformed to the surface of the facial template through $F_{\text{dfm}}$, and $T_{\text{rfl}}$ return the corresponding BRDF attributes $\mathcal{B}(\mathbf{x})=[b_{\text{a}}(\mathbf{x}), b_{\text{s}}(\mathbf{x}), b_{\text{r}}(\mathbf{x})]\in\mathbb{R}^8$. 
Subsequently, the specific reflectance attributes for each individual are determined by learning the BRDF offsets with $F_{\text{ofs}}$.
To capture the unique reflectance characteristics associated with individual identities, we incorporate a reflectance latent code $\mathbf{z}_{\text{rfl}}\in\mathbb{R}^{128}$.
Moreover, we adjust the offset using a weighting parameter $w$, setting it to a low value of $w=0.05$. This choice reflects our preference for the template network to primarily capture reflectance properties, minimizing dependence on the BRDF offset network. 
To simulate environment light, we apply the spherical harmonics as in Eq.~\eqref{eq:sh} and integrate the lights as in~\cite{zheng2023neuface}.

After obtaining the geometry, reflectance, and environment light information, we calculate the final color (output radiance) as Eq.~(\ref{eq:int_L_o}-\ref{eq:sh}) and integrate it by volume rendering.
As shown in Fig.~\ref{fig:template}, a general facial template can be obtained and robustly decomposed into geometry, diffuse, and specular attributes.


\noindent\textbf{Albedo gradient predictor.} Considering the wide range of skin color differences among various ethnic groups, simply relying on the absolute values of albedo on the facial template does not adequately convey the prior knowledge of faces. 
We observe that the continuity of facial albedo variations across 3D positions exhibits certain regularities, such as drastic changes around the edges of the lips and the eyes. Thus, we introduce the concept of \textbf{albedo gradient} as an additional attribute for the facial template. Specifically, during the training phase of Stage 1, we calculate the gradient of the albedo values $\nabla \mathbf{b}_{\text{a}}(\mathbf{x}) \in\mathbb{R}^3$ from the fitted neural field with respect to changes in 3D coordinates $\mathbf{x}\in\mathbb{R}^3$. This gradient plot effectively outlines facial contours like facial sketches. Following this, we use it to supervise the training of an additional albedo gradient predictor branch $\mathcal{G}(\mathbf{x})$ within the reflectance template, guiding the model to pay attention to the patterns of albedo changes. During training, once the point $F_\text{dfm}(\mathbf{x})$ in the template space is obtained, it serves as the input for $\mathcal{G}$ and the output target is $\nabla \mathbf{b}_{\text{a}}(\mathbf{x})$ in the observation space. In the end, the well-trained albedo gradient predictor provides more comprehensive information and precise guidance for single-person refinement in Stage 2.

\subsection{Individual decomposition under sparse views}
\label{subsec:stage2}
Taking the facial template learned from Stage 1 as prior, we perform a fine-detailed decomposition from sparse-view images of a single face in Stage 2.

\noindent\textbf{Geometry displacement and BRDF offset.} The geometry template $T_{\text{geo}}$ only represents a rough shape with shared features of multiple human faces. Additionally, the deformation net $F_{\text{dfm}}$ transforms the points on the surface of the facial template to the surface of a specific individual's face. However, the transformation is constrained, making it challenging to create topological features beyond the template domain and capture high-frequency details. Consequently,  we incorporate a displacement network $F_{\text{dis}}$ in Stage 2 to modify the SDF obtained from the geometry template for a more accurate reconstruction. 
Building on Eq.~(\ref{eq:g-temp}), we incorporate the offset term as follows: $\mathcal{S}(\mathbf{x})=T_{\text{geo}}(F_{\text{dfm}}(\mathbf{x}, \mathbf{z}_{\text{geo}}))+F_{\text{dis}}(\mathbf{x}).$

Based on the reflectance template $T_{\text{rfl}}$ learned from Stage 1, we calculate the offset value of BRDF through $F_{\text{ofs}}$, similar to Stage 1, to obtain the specific BRDF attributes for each individual. To address the distinct diversity between diffuse albedo $b_a$, specular $b_s$ and roughness $b_r$, we add a learnable weight term  $\mathbf{W}(\mathbf{x})=[\mathbf{w}_\text{a}, \mathbf{w}_\text{s}, \mathbf{w}_\text{r}]$ to control the contribution of each BRDF attribute from the reflectance template $T_{\text{rfl}}$, where $\mathbf{w}_\text{a}\in\mathbb{R}^3, \mathbf{w}_\text{s}\in\mathbb{R}^4, \mathbf{w}_\text{r}\in\mathbb{R}$ are the weights for diffuse albedo, specular, and roughness, respectively. Thus, we obtain the BRDF attributes as:
\begin{equation}
    \mathcal{B}_{\text{st2}}(\mathbf{x})=\mathbf{W}(\mathbf{x})\cdot T_{\text{rfl}}[F_{\text{dfm}}(\mathbf{x}, \mathbf{z}_{\text{geo}})] + F_{\text{ofs}}(\mathbf{x}).
\end{equation}
Considering that the diffuse albedo attribute has a large variation between different people, we initialize $\mathbf{w}_\text{a}$ with relatively low values.


\noindent\textbf{Attribute synergy.} 
In our framework, we decompose facial attributes into two components: geometry and reflectance. These components are not isolated but can enhance each other cooperatively. Sometimes, distinguishing between geometry and reflectance can be challenging, leading to inaccuracies in the decomposition process. 
For example, a dark region in a facial image might result from sunken areas or pigmentation spots on the face.
Conventional neural rendering techniques rely solely on color information from images, making it hard to distinguish such ambiguous scenes, often leading to incorrect reconstruction. Since our method can decouple rough geometry and reflectance attributes in Stage 1, we leverage the prior from the reflectance template to facilitate geometry inference, and vice versa. 
Specifically, we obtain the displacement $\mathcal{D}(\mathbf{x})$ as follows:
\begin{equation}
    \mathcal{D}(\mathbf{x})=F_{\text{dis}}(\mathbf{x},\text{Concat}(\mathbf{f}_{\text{def}},\mathbf{f}_{\text{geo}},\mathbf{f}_{\text{rfl}},\mathcal{G}(F_\text{dfm}(\mathbf{x})))),
\end{equation}
where $\mathbf{f}_{\text{dfm}}\in\mathbb{R}^{128}$, $\mathbf{f}_{\text{geo}}\in\mathbb{R}^{128}$, and $\mathbf{f}_{\text{rfl}}\in\mathbb{R}^{128}$ are extracted features from the output of the final hidden layer of $F_{\text{dfm}}$, $T_{\text{geo}}$, and $T_{\text{rfl}}$, respectively. $\mathcal{G}(F_\text{dfm}(\mathbf{x}))$ is the output from the albedo gradient predictor. We also extract another feature vector $\mathbf{f}_{\text{dis}}\in\mathbb{R}^{64}$ from $F_{\text{dis}}$ to provide more geometry information to the BRDF offset network $F_{\text{ofs}}$. 
Similarly, we calculate the BRDF offset $\mathcal{B}_{\text{ofs}}(\mathbf{x})$ as: 
\begin{equation}
    \mathcal{B}_{\text{ofs}}(\mathbf{x})=F_{\text{ofs}}(\mathbf{x},\text{Concat}(\mathbf{f}_{\text{def}},\mathbf{f}_{\text{geo}},\mathbf{f}_{\text{dis}},\mathbf{f}_{\text{rfl}},\mathcal{G}(F_\text{dfm}(\mathbf{x})))).
\end{equation}
Our ablation study validates the effectiveness of this attribute synergy module in the accuracy of both geometry reconstruction and rendering results.

\begin{figure}[htbp]
    \centering

\includegraphics[width=0.9\linewidth,trim={0cm 3cm 0cm 0cm}]{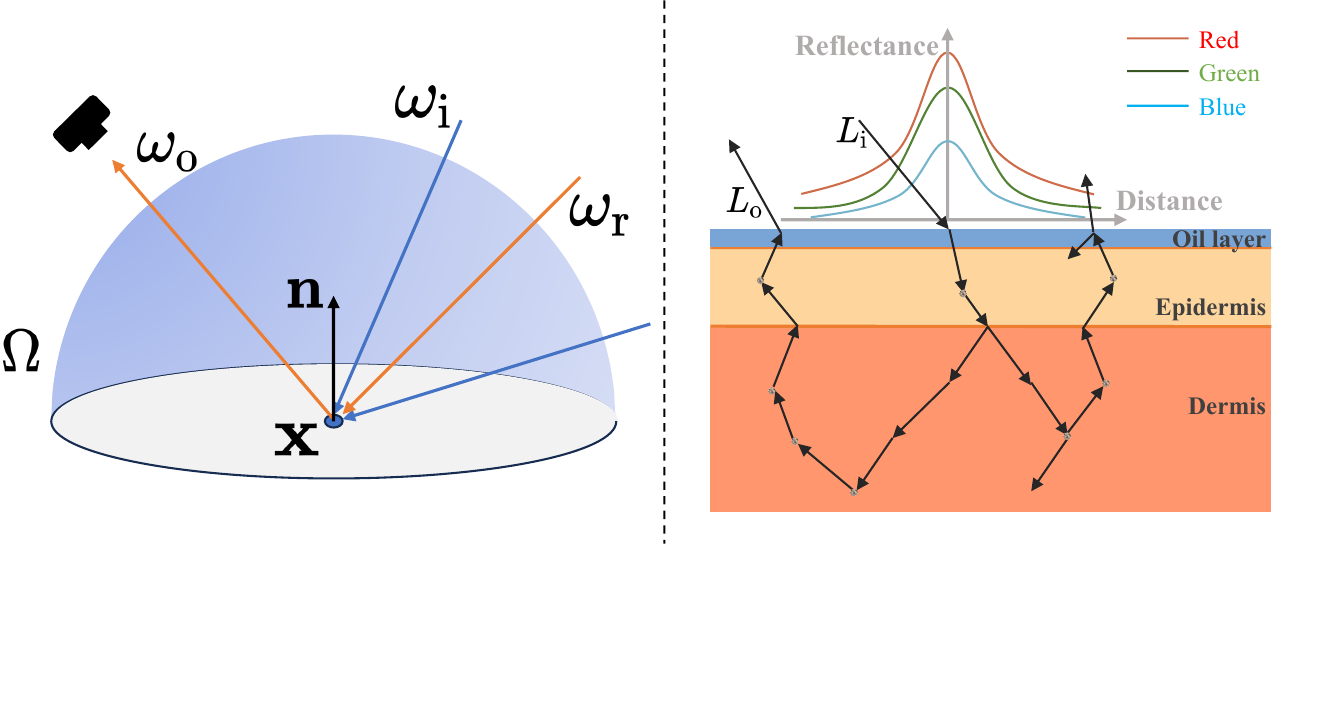}
    \caption{\textbf{Left:} BRDF rendering, where light reflects at the incident point. \textbf{Right:} Subsurface scattering on facial skin, showing the offset between the incident and outgoing points.}
    \label{fig:render_process}
\end{figure}

\noindent\textbf{Subsurface scattering offset.} Human facial skin exhibits a complexity far beyond that of non-biological materials like metal and plastic. Building on insights from previous research~\cite{donner2005light}, rendering realistic faces necessitates accounting for subsurface scattering effects. As illustrated in Fig.~\ref{fig:render_process}, this phenomenon can be modeled using a three-layer approach, where specular reflections occur directly at the point of light incidence on the surface, while diffuse reflections involve light penetrating the skin, traversing three layers, undergoing multiple subsurface reflections, and then emerging from the skin surface slightly away from the point of incidence. This effect, known as subsurface scattering, is crucial for realistic skin rendering and can be mathematically represented as $L_{\text{o}}(\mathbf{x}_{\text{o}}, \omega_{\text{o}})=\int_{A} \int_{\Omega } S(\mathbf{x}_{\text{o}}, \boldsymbol{\omega}_{\text{o}}, \mathbf{x}_{\text{i}}, \boldsymbol{\omega}_{\text{i}}) L_{\text{i}}(\mathbf{x}_{\text{i}}, \boldsymbol{\omega}_{\text{i}})(\boldsymbol{\omega}_{\text{i}}\cdot\mathbf{n}) \mathrm{d} \boldsymbol{\omega}_{\text{i}} \mathrm{d} A$, where $\mathbf{x}_\text{i}$ and $ \mathbf{x}_\text{o}$ are the positions of the light incident and output points, respectively, $A$ is a small area around $\mathbf{x}_{\text{o}}$, and $S$ is the scattering evaluation. In practice, the BRDF mainly accounts for light that scatters back at the point where it hits a surface, but simulating surface scattering accurately is computationally expensive. To improve face reconstruction quality, we integrate subsurface scattering into the BRDF's diffuse component as an offset. We simplify the process by assuming uniform subsurface scattering properties within a small area and utilize scattering profiles to approximate this effect. As illustrated in Fig.~\ref{fig:render_process}, these profiles show how the reflectance due to subsurface scattering changes with the distance between where light enters and exits the skin.

As ~\cite{lauritzen2007gpu} suggests, one can approximate each RGB channel of the scattering profiles using a combination of six Gaussian functions. We employ the weights of these Gaussian functions to form a feature vector $\mathbf{f}^{\mathbf{x}}_{\text{sp}}\in\mathbb{R}^{18}$, which represents the scattering profile of the point $\mathbf{x}$. Consequently, we calculate the offset of the diffuse term as:
\begin{equation}
    F_{\text{ss}}(\mathbf{x})=v\hat{F}_{\text{sp}}(\mathbf{x},\mathbf{f}^{\mathbf{x}}_{\text{sp}})\hat{L}_{\text{i}}(\mathbf{x},\mathcal{E}(\mathbf{x})),
\end{equation}
where $v$ is the weight of scattering offset, $\hat{F}_{\text{sp}}$ is a network to integrate scattering reflectance, $\hat{L}_{\text{i}}$ is a network to integrate the incoming light in a small region around $\mathbf{x}$, and $\mathcal{E}(\mathbf{x})\in\mathbb{R}^3$ is a learnable vector that defines the size of the integration domain. The values of $\mathbf{f}^{\mathbf{x}}_{\text{sp}}$ and $\mathcal{E}(\mathbf{x})$ are obtained through a parameter network. We refer readers to the supplementary for more details of the network architecture.

\subsection{Training losses}
For both stages, we consider reconstruction losses including the pixel-wise color loss $\mathcal{L}_{\text{col}}=\|\mathbf{C}-\mathbf{C}_{\text{gt}}\| _{1}$ and the eikonal loss $\mathcal{L}_{\text{eik}}=(\|\nabla \mathcal{S}\| _{2} -1)^2$. Similar to NeuFace~\cite{zheng2023neuface}, we incorporate a regularization term $\mathcal{L}_{\text{light}}$ to assume a nearly white environment light and penalize the specular energy by $\mathcal{L}_{\text{spec}}$. Additionally, we add $\mathcal{L}_{\text{code}}=\|\mathbf{z}_{\text{geo}}\|_2+\|\mathbf{z}_{\text{rfl}}\|_2$ to regularize the latent codes. To avoid a large variation between the predict reflectance attributes and the template during training, we apply a constraint to BRDF offset as $\mathcal{L}_{\text{ofs}}=\|\mathcal{B}_{\text{ofs}}(\mathbf{x})\|_1$. 
Additionally, the albedo gradient predictor employs $\mathcal{L}_{\text{g}}$, an L1 loss, which is supervised by the computed albedo gradient.
As a result, we formulate the loss function in Stage 1 as follows:
\begin{equation}
    \mathcal{L}_{\text{st1}}=\mathcal{L}_{\text{col} }+\mathcal{L}_{\text{eik} }+\mathcal{L}_{\text{light} }+\mathcal{L}_{\text{spec}}+\mathcal{L}_{\text{ofs}}+\mathcal{L}_{\text{code}}+\mathcal{L}_{\text{g}}.
\end{equation}
For the second stage, we add additional displacement loss $\mathcal{L}_{\text{dis}}$ and scatter offset loss $\mathcal{L}_{\text{ss}}$: $\mathcal{L}_{\text{st2}}=\mathcal{L}_{\text{st1}}+\mathcal{L}_{\text{dis}}+\mathcal{L}_{\text{ss} }$, where $\mathcal{L}_{\text{dis}}=\|\mathcal{D}(\mathbf{x})\|_1$ and $\mathcal{L}_{\text{ss}}=\|F_{\text{ss}}(\mathbf{x})\|_1$ are regularization losses aimed at constraining high values of the displacement and the subsurface scattering offset.


\begin{table*}[htbp]
\centering
    \resizebox{0.95\textwidth}{!}{
        \begin{tabular}{c|c|*{4}{c}|*{4}{c}|*{4}{c}}
        \toprule
                        \multirow{2}{*}{Method} & \multirow{2}{*}{Relight}
                        & \multicolumn{4}{c|}{\textbf{3 views}} 
                        & \multicolumn{4}{c|}{\textbf{5 views}} 
                        & \multicolumn{4}{c}{\textbf{10 views}}  
                        \\
                        & & PSNR\(\uparrow\)\ & SSIM\(\uparrow\)\ & LPIPS\(\downarrow\)\ &  CD\(\downarrow\)\ 
                        & PSNR\(\uparrow\)\ & SSIM\(\uparrow\)\ & LPIPS\(\downarrow\)\ &  CD\(\downarrow\)\ 
                        & PSNR\(\uparrow\)\ & SSIM\(\uparrow\)\ & LPIPS\(\downarrow\)\ &  CD\(\downarrow\)\  \\
        \midrule
        \midrule
        VolSDF~\cite{yariv2021volume} & \xmark  & 22.95 & 0.868 & 6.92 & 18.28
        & 25.45 & 0.897 & 6.09 & \cellcolor{orange!25}7.40 
        & 27.34 & 0.910 & 5.18 & \cellcolor{red!25}\textbf{5.25}  \\
        DeformHead~\cite{xu2023deformable}     & \xmark & \cellcolor{red!25}\textbf{26.02} & \cellcolor{orange!25}0.886 & \cellcolor{orange!25}5.76 & \cellcolor{orange!25}9.60
        & \cellcolor{orange!25}26.41 & 0.879 & 4.83 & 8.44
        & \cellcolor{orange!25}27.35 & 0.891 & 5.20 & \cellcolor{orange!25}5.42 \\
        \midrule
        PhySG~\cite{zhang2021physg} & \cmark    & 22.03 & 0.863 & 7.81 & 70.87
        & 22.69 & 0.872 & 7.05 & 63.20 
        & 23.11 & 0.876 & 7.12 & 55.7 \\
        TensoIR~\cite{jin2023tensoir} & \cmark & 22.35 & 0.645 & 7.14 & 28.60
        & 24.61 &  0.704 & 5.86 & 17.10
        & 26.14 & 0.769 & 4.73 & 10.10 \\
        NeuFace~\cite{zheng2023neuface} & \cmark     & 23.54 & 0.876 & 5.78 & 15.74 
        & 25.81 & \cellcolor{orange!25}0.904 & \cellcolor{orange!25}4.75 & 8.62 
        & 27.19 & \cellcolor{orange!25}0.914 & \cellcolor{orange!25}4.03 & 6.40 \\
         Ours  & \cmark    & \cellcolor{orange!25}25.54 & \cellcolor{red!25}\textbf{0.902} & \cellcolor{red!25}\textbf{4.46} & \cellcolor{red!25}\textbf{9.27}
        & \cellcolor{red!25}\textbf{26.73}  & \cellcolor{red!25}\textbf{0.911} & \cellcolor{red!25}\textbf{3.83}	& \cellcolor{red!25}\textbf{7.00} 
        & \cellcolor{red!25}\textbf{27.64} & \cellcolor{red!25}\textbf{0.920} & \cellcolor{red!25}\textbf{3.86} & 5.53 \\
        \bottomrule
    \end{tabular}}
    \caption{Evaluation. The LPIPS and CD metrics are measured on scales of $10^{-2}$ and $10^{-3}$, respectively. The best and second-best results are highlighted in red and orange, respectively.}
    \vspace{-5pt}
    \label{tab:table_comparison}
\end{table*}

\section{Experimental results}
\label{sec:experiments}

\subsection{Experimental setup}
\textbf{Dataset and metrics.} We evaluate our method on a large human face dataset \textit{Facescape}~\cite{yang2020facescape}.
Specifically, we utilize authorized data from approximately 30 individuals, 4 of whom are from the publishable list. To protect the privacy of other identities that are not on the publishable list, we employed mosaic technology in the eyes region. 
To validate the generalization performance of our method, we additionally use dataset \textit{H3DS}~\cite{ramon2021h3d} in the supplementary material.
We assess the quality of image rendering using the PSNR, SSIM, and LPIPS~\cite{zhang2018unreasonable} metrics, and the Chamfer distance (CD) for evaluating the accuracy of geometry reconstruction. We compare our method with two 3D reconstruction methods VolSDF~\cite{yariv2021volume}, DeformHead~\cite{xu2023deformable} and three BRDF decomposition methods PhySG~\cite{zhang2021physg}, TensoIR~\cite{jin2023tensoir}, and NeuFace~\cite{zheng2023neuface}.

\noindent\textbf{Implementation details.} \textit{Facescape} has over 50 scans of human heads along with camera parameters and meshes. We preprocess the images to obtain face masks using ImFace~\cite{zheng2022imface} tools. For Stage 1 training, we select 10 subjects, each with 10 views. For Stage 2, we use an additional 20 subjects and choose 3, 5, and 10 views from the available 50+ views as the training set, with another 10 views reserved for testing. The 10-view experiments specifically evaluate performance in a low-view setting. All test data is independent of the training set and is used consistently across all methods for comparison.
We employ the Adam optimizer with an initial learning rate $1\times 10^{-4}$ to train both stages for 3000 epochs. Additionally, we add a calibration module as ~\cite{zheng2023neuface} to mitigate the exposure differences between various views and apply this calibration to all the methods for a fair comparison. For more implementation details, please refer to the supplementary material.

\noindent\subsection{Evaluation \& comparison}
\begin{figure}[htbp]
    \centering
    \makebox[0.02\linewidth]{\rotatebox{90}{\hspace{15pt}\footnotesize VolSDF}}\hspace{5pt}
    \includegraphics[width=0.19\linewidth]{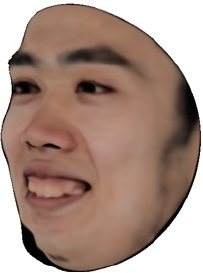}\hspace{5pt}
    \includegraphics[width=0.19\linewidth]{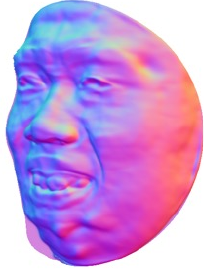}\hspace{5pt}
    \includegraphics[width=0.19\linewidth]{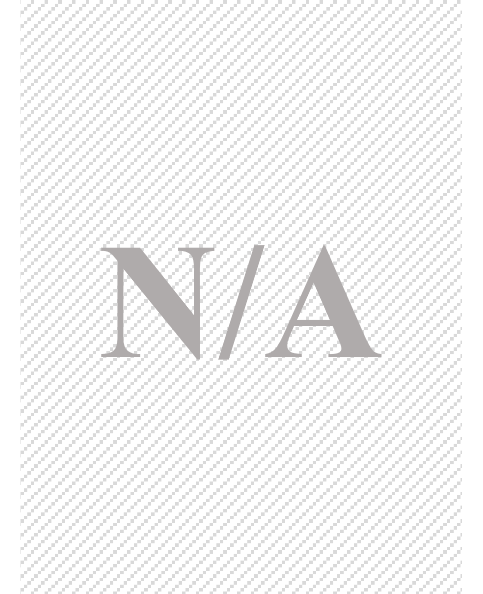}\hspace{5pt}
    \includegraphics[width=0.19\linewidth]{low_res_vis/trimed_comparison/NA.png}
    \\
    \makebox[0.02\linewidth]{\rotatebox{90}{\hspace{12pt}\footnotesize DeformHead}}\hspace{5pt}
    \includegraphics[width=0.19\linewidth]{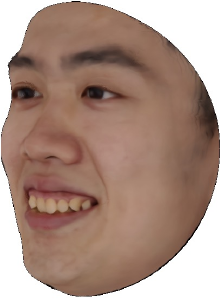}\hspace{5pt}
    \includegraphics[width=0.19\linewidth]{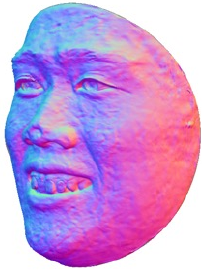}\hspace{5pt}
    \includegraphics[width=0.19\linewidth]{low_res_vis/trimed_comparison/NA.png}\hspace{5pt}
    \includegraphics[width=0.19\linewidth]{low_res_vis/trimed_comparison/NA.png}
    \\
    \makebox[0.02\linewidth]{\rotatebox{90}{\hspace{17pt}\footnotesize PhySG}}\hspace{5pt}
    \includegraphics[width=0.19\linewidth]{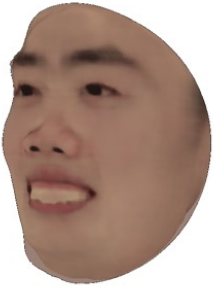}\hspace{5pt}
    \includegraphics[width=0.19\linewidth]{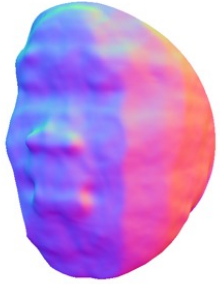}\hspace{5pt}
    \includegraphics[width=0.19\linewidth]{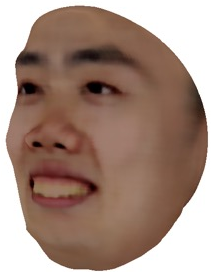}\hspace{5pt}
    \includegraphics[width=0.19\linewidth]{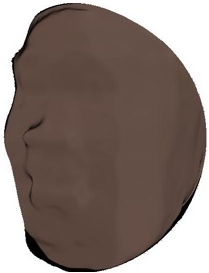}
    \\
    \makebox[0.02\linewidth]{\rotatebox{90}{\hspace{15pt}\footnotesize TensoIR}}\hspace{5pt}
    \includegraphics[width=0.19\linewidth]{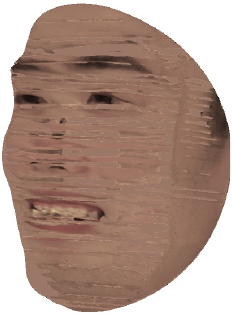}\hspace{5pt}
    \includegraphics[width=0.19\linewidth]{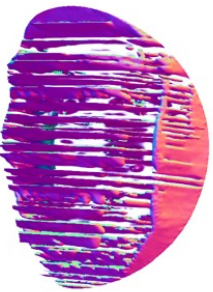}\hspace{5pt}
    \includegraphics[width=0.19\linewidth]{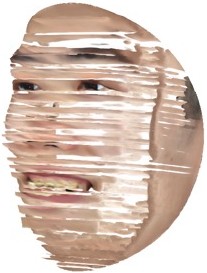}\hspace{5pt}
    \includegraphics[width=0.19\linewidth]{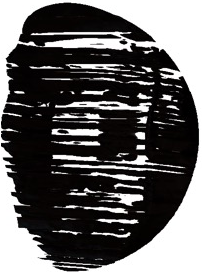}
    \\
    \makebox[0.02\linewidth]{\rotatebox{90}{\hspace{15pt}\footnotesize NeuFace}}\hspace{5pt}
    \includegraphics[width=0.19\linewidth]{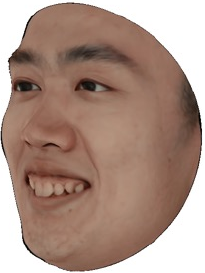}\hspace{5pt}
    \includegraphics[width=0.19\linewidth]{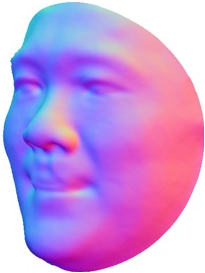}\hspace{5pt}
    \includegraphics[width=0.19\linewidth]{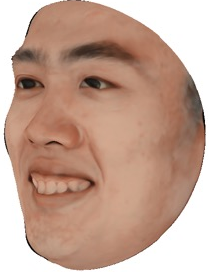}\hspace{5pt}
    \includegraphics[width=0.19\linewidth]{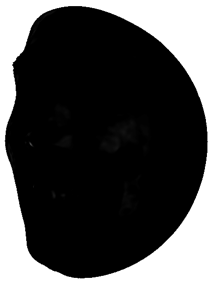}
    \\
    \makebox[0.02\linewidth]{\rotatebox{90}{\hspace{18pt}\footnotesize \textbf{Ours}}}\hspace{5pt}
    \includegraphics[width=0.19\linewidth]{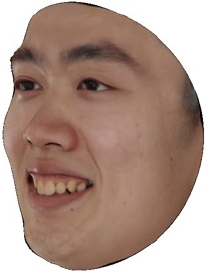}\hspace{5pt}
    \includegraphics[width=0.19\linewidth]{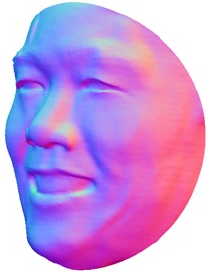}\hspace{5pt}
    \includegraphics[width=0.19\linewidth]{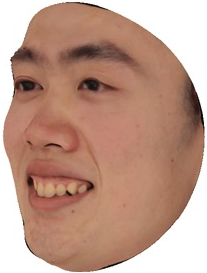}\hspace{5pt}
    \includegraphics[width=0.19\linewidth]{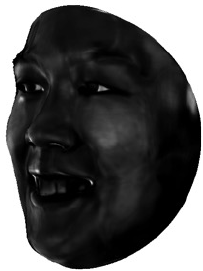}
    \\
    \makebox[0.02\linewidth]{}\hspace{5pt}
    \makebox[0.19\linewidth]{\footnotesize\textbf{Rendering}}\hspace{5pt}
    \makebox[0.19\linewidth]{\footnotesize\textbf{Normal}}\hspace{5pt}
    \makebox[0.19\linewidth]{\footnotesize\textbf{Diffuse}}\hspace{5pt}
    \makebox[0.19\linewidth]{\footnotesize\textbf{Specular}}\hspace{5pt}
    \caption{\textbf{Visualization of 3-view reconstruction results.} VolSDF and DeformHead do not incorporate reflectance attributes, resulting in the absence of corresponding outcomes.}
    \label{fig:comparison}
\end{figure}

To corroborate the effectiveness of our proposed framework in the reconstruction of appearance and geometry, we conducted a comprehensive comparison with state-of-the-art methods. These methods are categorized into two groups: geometry-oriented methods, which include VolSDF~\cite{yariv2021volume} and DeformHead~\cite{xu2023deformable}; and decomposition-based methods, including PhySG~\cite{zhang2021physg}, TensoIR~\cite{jin2023tensoir}, NeuFace~\cite{zheng2023neuface}, and our method \textbf{SFDM}. As shown in Tab.~\ref{tab:table_comparison}, we enumerate the quantitative performance of different methods in the evaluation metrics. To assess the performance capability under sparse-view images, we conducted tests on 20 subjects with 3, 5, and 10 views, respectively. The visual results of the reconstruction from 3 views are presented in Fig.~\ref{fig:comparison}, showcasing the synthesized rendering images from a novel viewpoint, the geometric reconstruction, and the decomposition of diffuse and specular attributes.

\noindent\textbf{Decomposition-based methods.} 
Our method exhibits superior performance in novel view rendering and geometry reconstruction compared to other reflectance decomposition methods. In contrast to PhySG and TensoIR, designed for common objects, our method significantly outperforms them across all metrics. This is due to the complexity of facial material reflectance, which requires specialized reflectance modeling. 
NeuFace, which is customized for facial skin, poses a challenge due to its non-robust decomposition of geometry and reflectance attributes. As shown in Fig.~\ref{fig:comparison}, it often happens in NeuFace that all reflectance is baked into albedo, but the specular term is entirely absent. This degenerate Lambertian model may reduce the realism of the synthesized rendering output, especially for elements such as teeth, eyes, or the oily layer on facial skin, which often exhibit specular reflection effects. In contrast, we are able to stably decompose the diffuse and specular properties, resulting in more realistic rendering effects. 
Additionally, we demonstrate higher accuracy in geometric reconstruction, whereas the quality of NeuFace's geometry deteriorates significantly as the number of views decreases.


\noindent\textbf{Geometry-oriented methods.} Our method demonstrates a comparable level of accuracy in geometry reconstruction compared to VolSDF and DeformHead. VolSDF exhibits poor 3D reconstruction results in sparse view settings. While DeformHead achieves a more detailed 3D reconstruction utilizing a geometry template, it is overly sensitive to geometry deformation, leading to noticeable indentations. Although the PSNR of our method is lower than that of DeformHead under 3 views, our LPIPS outperforms it, indicating more photorealistic synthesis results. Failure instances in reconstruction have been observed with VolSDF and DeformHead, with failure rates of 21.7\%  and 6.7\%, respectively. We exclude these failures from the comparison.


\begin{figure}[htbp]
    \centering
    
    \resizebox{1.0\linewidth}{!}{
    \small
    \begin{tabular}{lc|*{4}{c}}
        \toprule
        Exp & Method & PSNR\(\uparrow\)\ & SSIM\(\uparrow\)\ & LPIPS\(\downarrow\)\ &  CD\(\downarrow\)\   \\
        \midrule
        \midrule
        1 & w/o facial template  & 24.27 & 0.857 & 5.59 & 17.22  \\
        2 & w/o albedo gradient  & \cellcolor{orange!25}25.13 & \cellcolor{orange!25}0.877 & \cellcolor{orange!25}4.66 & 8.83  \\
        \midrule
        3 & w/o attribute synergy  & 24.79 & 0.876 & 4.90 & 8.80  \\
        4 & w/o scattering offset  & 24.60 & 0.874 & 5.73 & \cellcolor{orange!25}8.61 \\
        \midrule
        5 & Ours &  \cellcolor{red!25}\textbf{25.65} & \cellcolor{red!25}\textbf{0.882} & \cellcolor{red!25}\textbf{4.45} & \cellcolor{red!25}\textbf{7.94}  \\
        \bottomrule
    \end{tabular}}
    \captionof{table}{Comparative evaluation metrics for ablation studies.}
    \vspace{-10pt}
    \label{tab:ablation}
\end{figure}

\begin{figure}[htbp]
    \centering
    \includegraphics[width=0.13\linewidth]{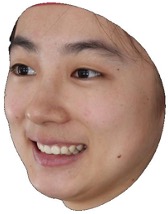}
    \includegraphics[width=0.13\linewidth]{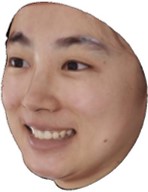}
    \makebox[0.003\linewidth]{}
    \includegraphics[width=0.13\linewidth]{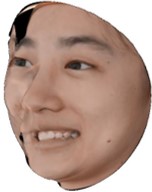}
    \makebox[0.003\linewidth]{}
    \includegraphics[width=0.13\linewidth]{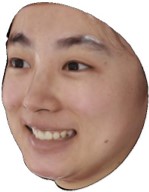}
    \includegraphics[width=0.13\linewidth]{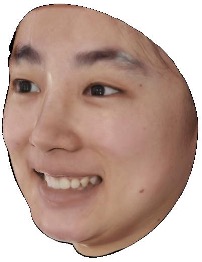}
    \includegraphics[width=0.13\linewidth]{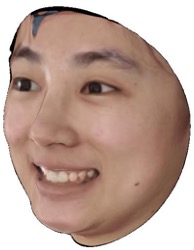}
    \makebox[0.03\linewidth]{}
    \\
    \includegraphics[width=0.13\linewidth]{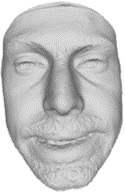}
    \includegraphics[width=0.13\linewidth]{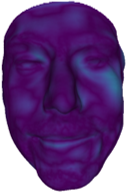}
    \includegraphics[width=0.162\linewidth]{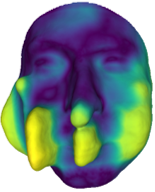}
    \includegraphics[width=0.13\linewidth]{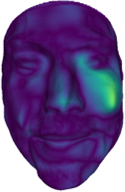}
    \includegraphics[width=0.13\linewidth]{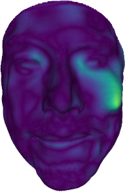}
    \includegraphics[width=0.13\linewidth]{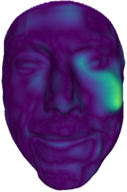}
    \includegraphics[width=0.03\linewidth]{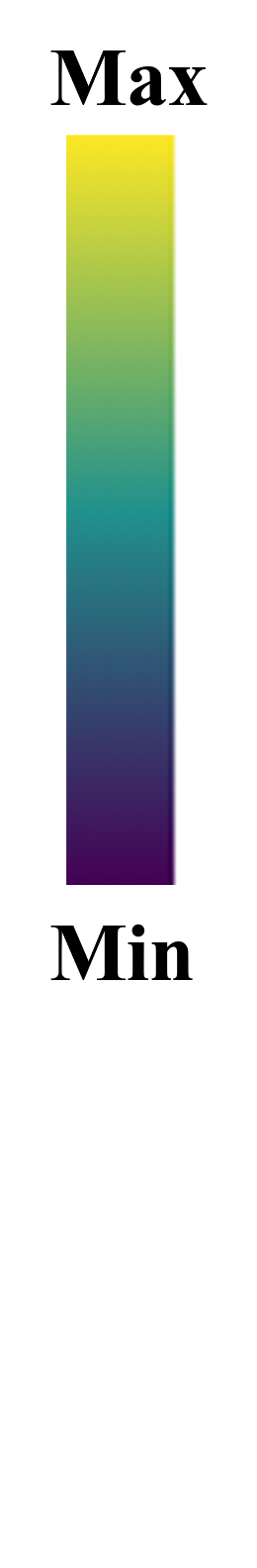}
    \\
    \makebox[0.13\linewidth]{\tiny GT}
    \makebox[0.13\linewidth]{\tiny Ours}
    \makebox[0.003\linewidth]{}
    \makebox[0.13\linewidth]{\tiny w/o $T_*$}
    \makebox[0.003\linewidth]{}
    \makebox[0.13\linewidth]{\tiny w/o $\mathcal{G}$}
    \makebox[0.13\linewidth]{\tiny w/o $\mathbf{f}_*$}
    \makebox[0.13\linewidth]{\tiny w/o $F_{\text{ss}}$}
    \makebox[0.03\linewidth]{}
    \vspace{-8pt}
    \caption{\textbf{Ablation visualization. Top:} Rendering. \textbf{Bottom:} Geometric errors.}
    \label{fig:ablation-cross}

\end{figure}

\subsection{Ablation studies}
We select a subset of 10 subjects from the dataset to test the effectiveness of the major components of \textbf{SFDM}. In Tab.~\ref{tab:ablation}, Exp.1-2 are results for the components in Stage 1, and Exp.3-4 cover the components in Stage 2. Exp.1-2 indicate that the facial template significantly influences the robustness of face decomposition, with the albedo gradient predictor further enriching the reconstruction quality by introducing additional facial prior knowledge. From Exp.3, we observe that the attribute synergy module appears to contribute to geometry decomposition, and its absence leads to a noticeable decrease in geometric accuracy. Based on Exp.4, the subsurface scatter offset module can enhance the realism of the synthesized images from novel views, which is suggested by improvements in the LPIPS metric. In Fig.~\ref{fig:ablation-cross}, columns 3-6 show the visualization results when omitting the facial template $T_*$, albedo gradient predictor $\mathcal{G}$, attribute synergy module $\mathbf{f}_*$, and subsurface scattering $F_{\text{ss}}$, respectively. The bottom row highlights the albedo gradient predictor's more significant impact on individuals who exhibit greater deviations from the facial template. In Stage 2, the subsurface scattering offset introduces additional details in light reflection on the skin, resulting in more photorealistic rendering results. Regarding the attribute synergy module, the exchange of information between geometry and reflectance enables a more accurate decomposition, as is evident in the details of geometry reconstruction.

    

\subsection{Reflectance decomposition analysis} 

\begin{figure}[htbp]
    \vspace{0pt}
    \centering
    \includegraphics[width=0.6\linewidth]{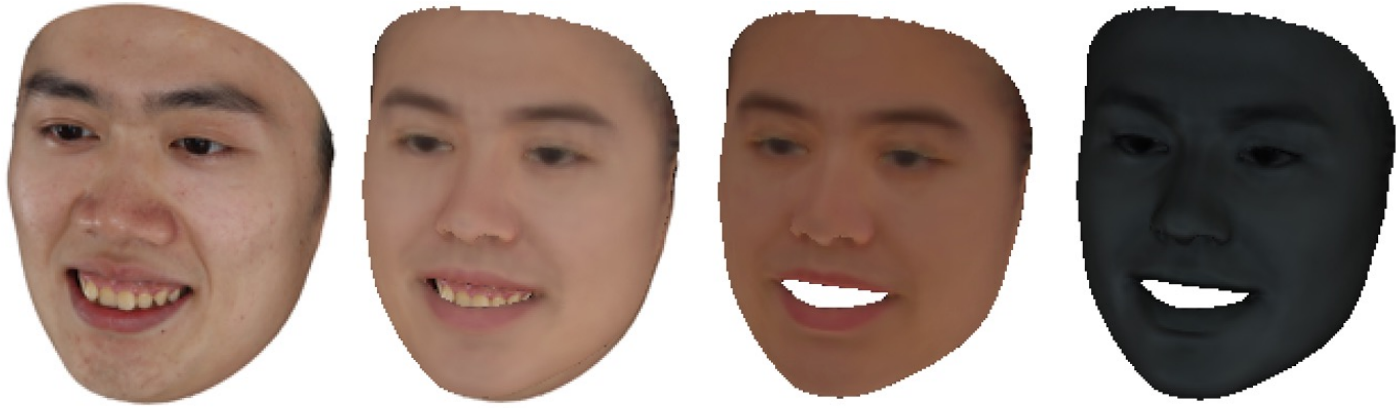}
    \\
    \hspace{5pt}
    \makebox[0.15\linewidth]{\footnotesize\textbf{GT}}\hspace{0pt}
    \makebox[0.15\linewidth]{\footnotesize\textbf{Render}}\hspace{0pt}
    \makebox[0.15\linewidth]{\footnotesize\textbf{Diffuse}}\hspace{0pt}
    \makebox[0.15\linewidth]{\footnotesize\textbf{Specular}}\hspace{5pt}
    \vspace{-5pt}
    \caption{Decomposition results of RefMM~\cite{han2023learning}.}
    \label{fig:refmm}
\end{figure}

Our method focuses on reflectance decomposition from multi-view facial images without any ground truth reflectance data, enabling realistic reconstruction of facial texture and highlights for novel view rendering. 
While our goal is not a physically accurate decomposition of skin reflectance, we aim to evaluate our decomposition results both visually and quantitatively. 
To this end, we use the single-view face reconstruction model, RefMM~\cite{han2023learning}, to fit and decompose images in the test set, generating a coarse diffuse and specular result as pseudo ground truth for evaluation. 
Tab.~\ref{tab:reflectance_cmp} shows that our decomposition accuracy generally outperforms that of NeuFace. 
The overall low SSIM values are primarily due to the coarse nature of the pseudo ground truth, as illustrated in Fig.~\ref{fig:refmm}. 
This reflects certain advantages and limitations of single-view decomposition methods. These methods are typically trained on extensive facial image datasets with specialized lighting conditions or pseudo reflectance ground truth, allowing them to robustly decompose faces from monocular images. However, because they rely heavily on a parameterized facial model and limited input with less 3D information, the reconstructed results lack many facial details, making the rendered images appear less realistic. As shown in Fig.~\ref{fig:refmm} and Fig.~\ref{fig:application}, we can observe these differences visually in the rendered results.


\begin{figure}[htbp]
    \centering
    
    \resizebox{0.6\linewidth}{!}{
    \small
    \begin{tabular}{c|cc}
        \toprule
         Method & Diffuse & Specular  \\
        \midrule
        \midrule
        NeuFace & \cellcolor{orange!25}0.726 & \cellcolor{orange!25}0.552 \\
        Ours & \cellcolor{red!25}\textbf{0.761} & \cellcolor{red!25}\textbf{0.692} \\
        \bottomrule
    \end{tabular}}
    \vspace{-5pt}
    \captionof{table}{Quantitative diffuse and specular results (SSIM\(\uparrow\)).}
    \vspace{-15pt}
    
    \label{tab:reflectance_cmp}
\end{figure}

\subsection{Applications}

\begin{figure}[htbp]
    \vspace{-5pt}
    \centering
    
    \makebox[0.39\linewidth]{}
    \makebox[0.12\linewidth]{\small Original}
    \makebox[0.01\linewidth]{}
    \colorbox{yellow!40}{\includegraphics[clip,width=0.10\linewidth]{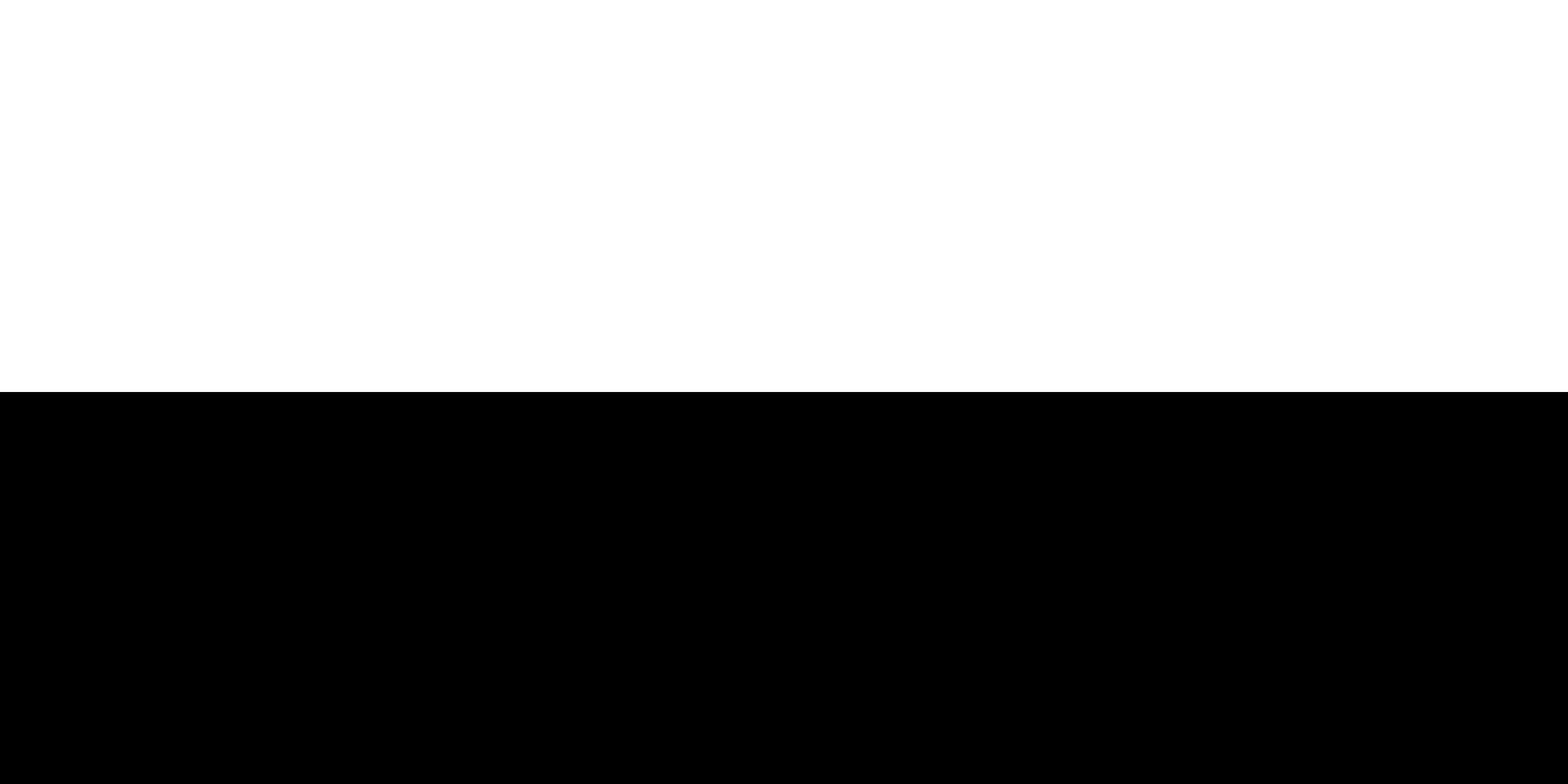}}
    \makebox[0.01\linewidth]{}
    \colorbox{yellow!40}{\includegraphics[clip,width=0.10\linewidth]{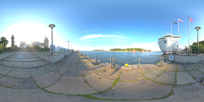}}
    \makebox[0.01\linewidth]{}
    \colorbox{yellow!40}{\includegraphics[clip,width=0.10\linewidth]{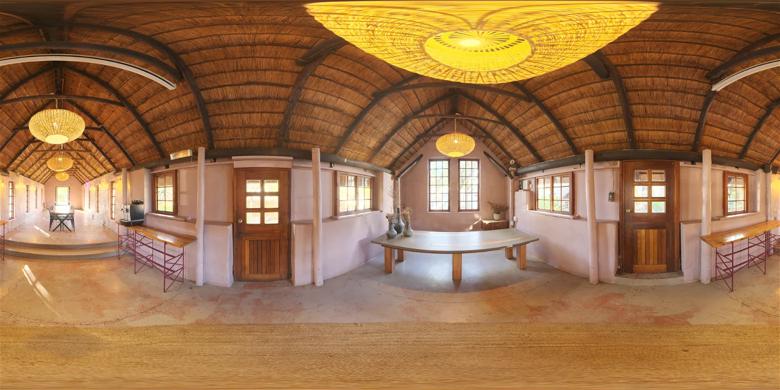}}
    \\
    \makebox[0.02\linewidth]{\rotatebox{90}{\hspace{10pt}\textbf{\footnotesize Raw}}}\hspace{5pt}
    \includegraphics[clip,trim={6cm, 6cm, 3cm, 12cm},width=0.13\linewidth]{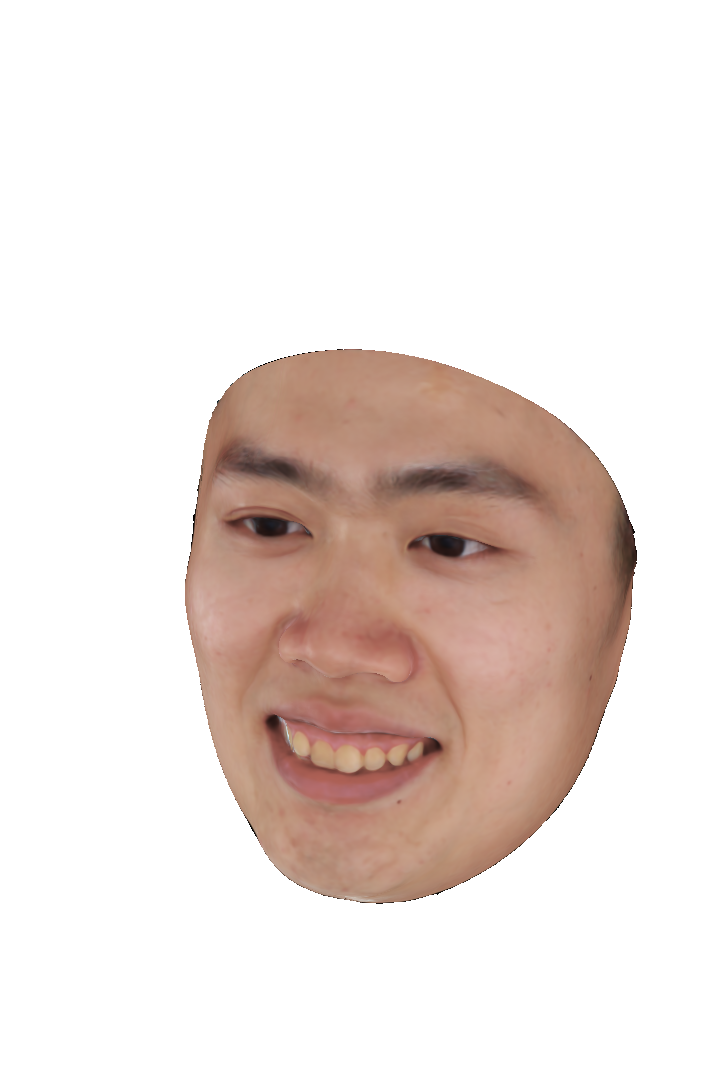}
    \includegraphics[clip,trim={6cm, 6cm, 3cm, 12cm},width=0.13\linewidth]{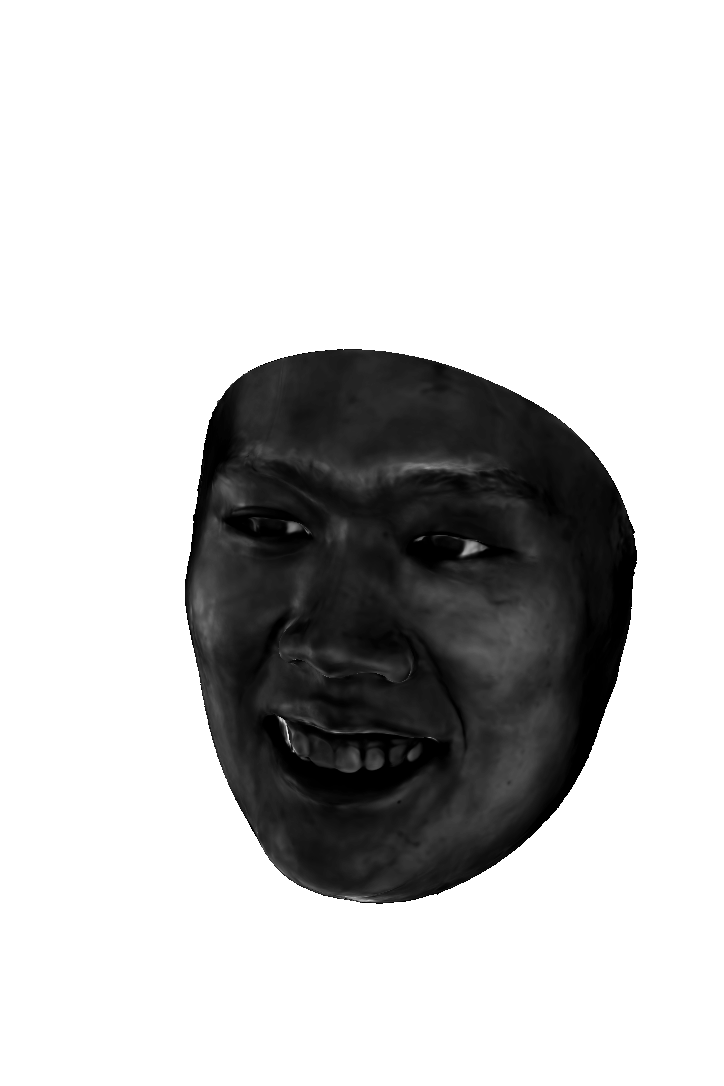}
    \hspace{5pt}
    \makebox[0.02\linewidth]{\rotatebox{90}{\hspace{5pt}\textbf{\footnotesize NeuFace}}}
    \includegraphics[clip,trim={5cm, 3cm, 3cm, 3cm},width=0.14\linewidth]{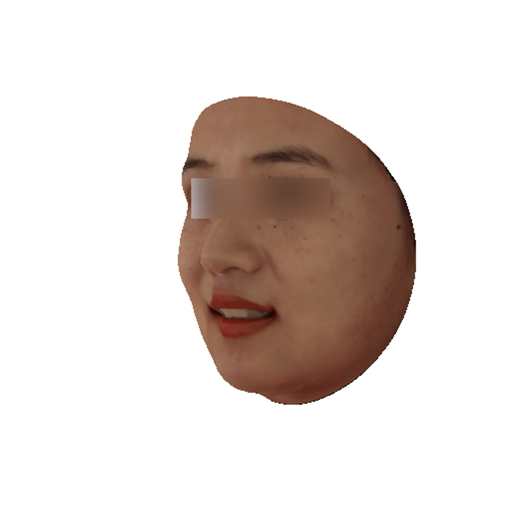}
    \includegraphics[clip,trim={5cm, 3cm, 3cm, 3cm},width=0.14\linewidth]{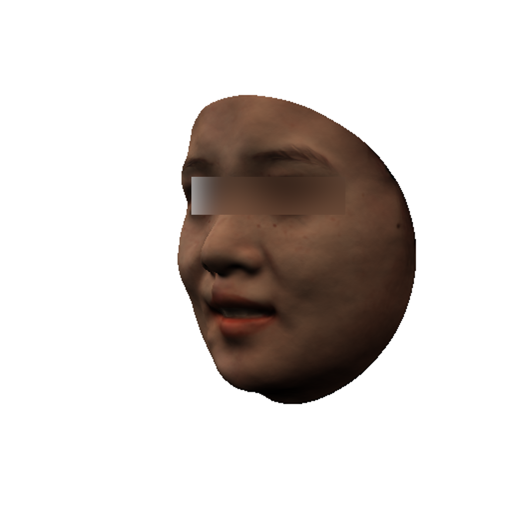}
    \includegraphics[clip,trim={5cm, 3cm, 3cm, 3cm},width=0.14\linewidth]{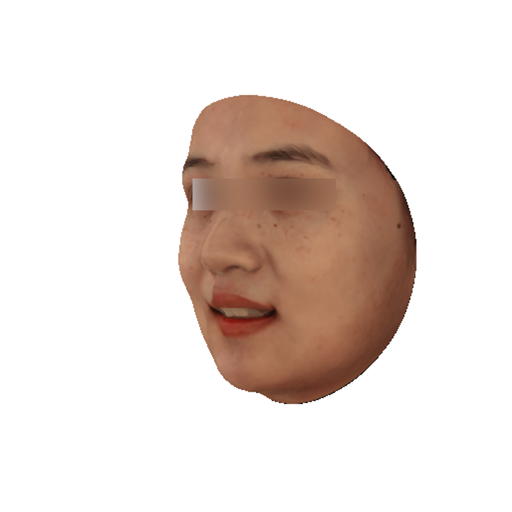}
    \includegraphics[clip,trim={5cm, 3cm, 3cm, 3cm},width=0.14\linewidth]{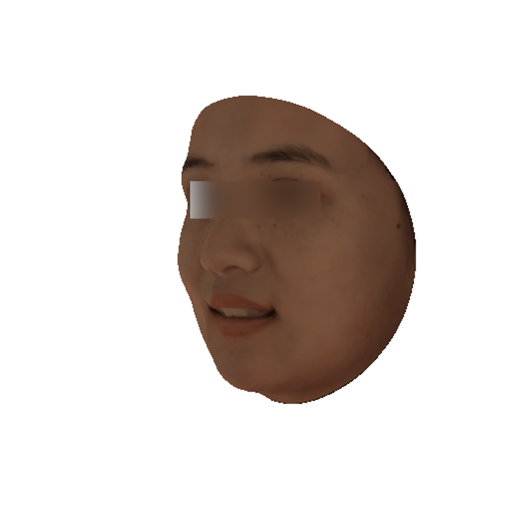}
    \\
    \vspace{-2pt}
    \makebox[0.02\linewidth]{\rotatebox{90}{\hspace{8pt}\textbf{\footnotesize Edited}}}\hspace{5pt}
    \includegraphics[clip,trim={6cm, 6cm, 3cm, 12cm},width=0.13\linewidth]{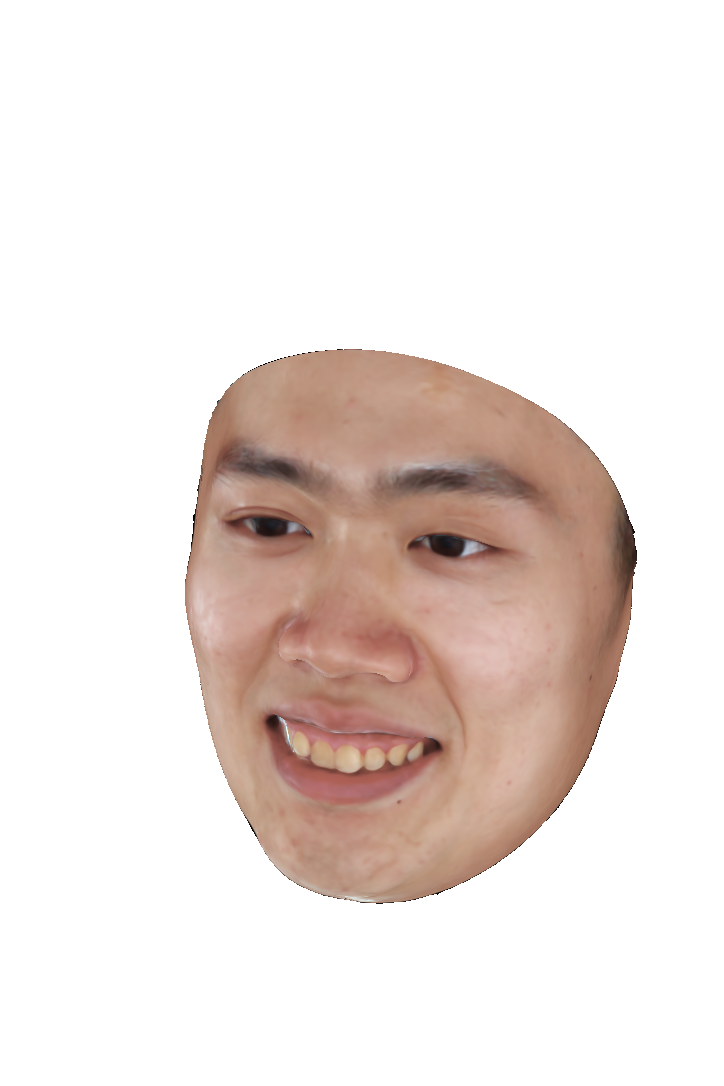}
    \includegraphics[clip,trim={6cm, 6cm, 3cm, 12cm},width=0.13\linewidth]{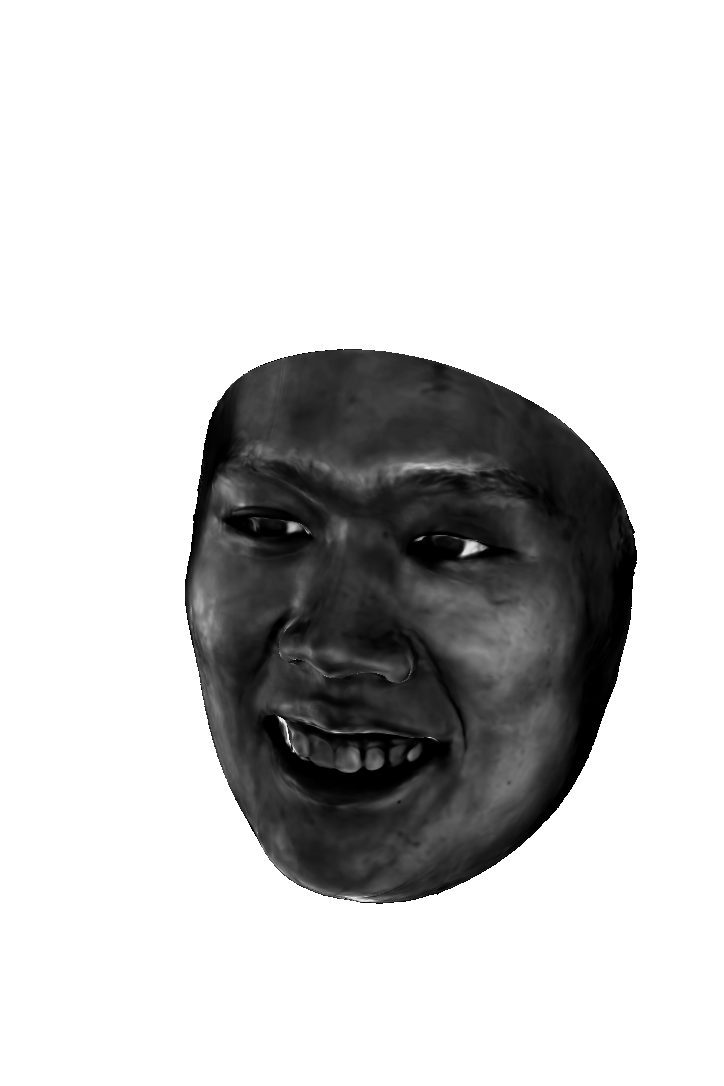}
    \hspace{5pt}
    \makebox[0.02\linewidth]{\rotatebox{90}{\hspace{13pt}\textbf{\footnotesize Ours}}}\hspace{0pt}
    \includegraphics[clip,trim={5cm, 3cm, 3cm, 3cm},width=0.14\linewidth]{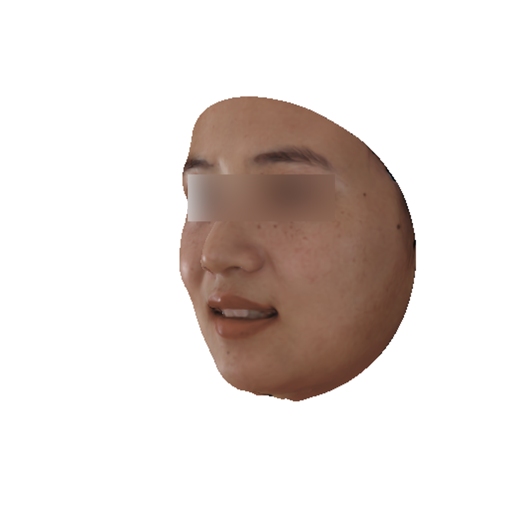}
    \includegraphics[clip,trim={5cm, 3cm, 3cm, 3cm},width=0.14\linewidth]{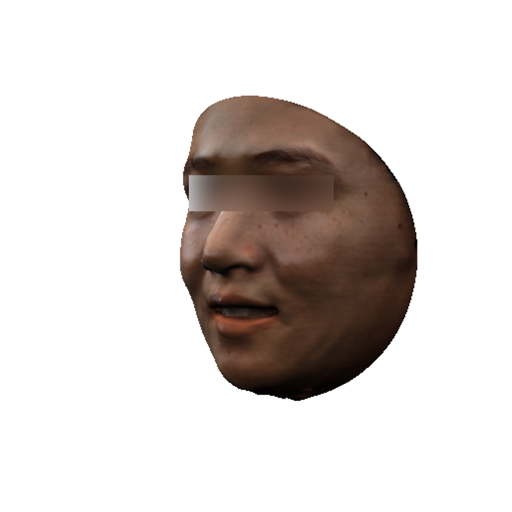}
    \includegraphics[clip,trim={5cm, 3cm, 3cm, 3cm},width=0.14\linewidth]{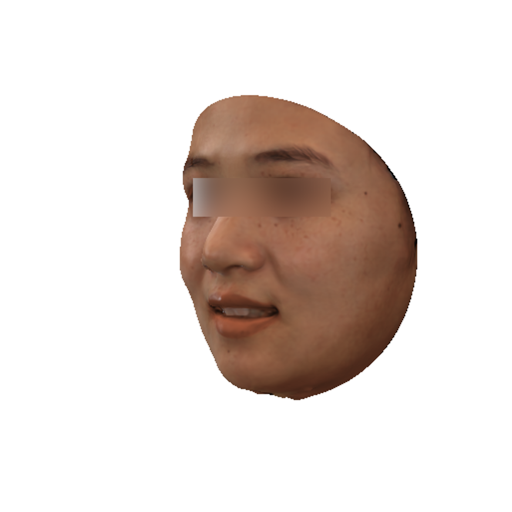}
    \includegraphics[clip,trim={5cm, 3cm, 3cm, 3cm},width=0.14\linewidth]{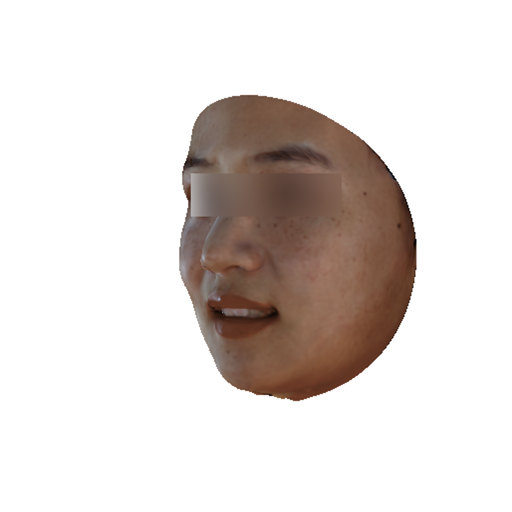}
    \\
    \makebox[0.02\linewidth]{\rotatebox{90}{\textbf{}}}\hspace{8pt}
    \makebox[0.22\linewidth]{\hspace{0pt}\footnotesize (a) Spec-edit}
    \makebox[0.25\linewidth]{}
    \makebox[0.19\linewidth]{\hspace{0pt}\footnotesize (b) Relighting}
    \makebox[0.22\linewidth]{}
    \\
    \vspace{-5pt}
    \caption{\textbf{Applications. (a)} Specular editing by increasing the shininess. \textbf{(b)} Relighting under different environments.}
    \label{fig:application}
    \vspace{-5pt}
\end{figure}

\noindent\textbf{Specular editing.} Our robust decomposition of diffuse and specular reflectances allows us to easily adjust the specular reflectance, resulting in the ability to make faces appear shinier or dimmer. The specular editing outcomes are showcased in Fig.~\ref{fig:application}. Specifically, we increase the specular weight by 1.5 times, while maintaining the overall specular distribution unchanged. As a result, the edited images can retain their photorealistic appearance.

\noindent\textbf{Relighting.} Decomposing reflectance and environment light facilitates the relighting of human faces without altering their inherent properties.  As shown in Fig.~\ref{fig:application}, we conducted a comparison of the relighting effects between our method and NeuFace in sparse view settings. Our method excels in decomposing specular reflectance, leading to more vibrant and realistic relighting outcomes under varying environmental lighting conditions~\cite{chen2020neural}.
\section{Conclusion}
We propose a novel face decomposition method capable of rendering and reconstructing 3D human faces from sparse-view images. Our framework first learns a facial template with both decomposed geometry and reflectance attributes from multi-view images of different individuals. Then we utilize the prior knowledge of the facial template and further consider the interaction between geometry and reflectance, as well as subsurface scattering effects, which enable our method to achieve robust face decomposition from sparse-view images.  Based on decomposed facial attributes, our method achieves superior results in the rendering and geometric reconstruction of 3D human faces and can be utilized for relighting and specular editing.


\noindent\textbf{Limitations \& future work.} As we focus on face decomposition, it is hard to extend to general objects directly with our method. 
In the future, we could involve techniques such as hash encoding~\cite{muller2022instant} to accelerate our method and leverage fine-detailed 3D prior or reflectance ground truth for a better decomposition effect.

\section*{Acknowledgment}
This work was supported in part by the Ministry of Education, Singapore, under its Academic Research Fund Grants (MOE-T2EP20220-0005 \& RT19/22) and the RIE2020 Industry Alignment Fund–Industry Collaboration Projects (IAF-ICP) Funding Initiative, as well as cash and in-kind contribution from the industry partner(s). J.Hu was supported by the National Natural Science Foundation of China under Grants 62402083.

{
    \small
    \bibliographystyle{ieeenat_fullname}
    \bibliography{main}
}


\end{document}